\documentclass{article} 
\usepackage{iclr2027_conference,times}
\iclrfinalcopy

\usepackage{amsmath,amsfonts,bm}

\def\eqref#1{equation~\ref{#1}}

\def\1{\bm{1}}

\DeclareMathAlphabet{\mathsfit}{\encodingdefault}{\sfdefault}{m}{sl}
\SetMathAlphabet{\mathsfit}{bold}{\encodingdefault}{\sfdefault}{bx}{n}

\usepackage{hyperref}
\usepackage{url}

\usepackage{amsmath}
\usepackage{multirow}
\usepackage{graphicx}
\usepackage{wrapfig}
\usepackage{algorithm}
\usepackage{algorithmic}
\usepackage{xcolor}
\usepackage{booktabs}
\usepackage{wrapfig}

\title{C2P-VAR: Continual and Compositional Personalization in Visual Autoregressive Models}

\author{%
  Junhao Li\textsuperscript{1}\thanks{Equal contribution.} \quad
  Xinhao Zhong\textsuperscript{1}\footnotemark[1] \quad
  Yi Sun\textsuperscript{1} \quad
  Yuxia Qiao\textsuperscript{4} \\ 
  \bfseries Bin Chen\textsuperscript{1,3}\thanks{Corresponding author.} \quad 
  Yaowei Wang\textsuperscript{1,3} \quad
  Shu-Tao Xia\textsuperscript{2} \\
  \textsuperscript{1}Harbin Institute of Technology, Shenzhen \\
  \textsuperscript{2}Tsinghua Shenzhen International Graduate School, Tsinghua University \\
  \textsuperscript{3}Peng Cheng Laboratory \\
  \textsuperscript{4}South China University of Technology \\
}

\begin{document}

\maketitle

\begin{abstract}

  Visual autoregressive (VAR) models have recently emerged as an efficient paradigm for text-to-image generation, yet their personalization capabilities remain largely limited to static, single-concept settings. In practice, users may continuously introduce new concepts and wish to compose multiple personalized concepts within a single image. Such scenarios pose two fundamental challenges: catastrophic forgetting during sequential personalization and feature interference during multi-concept composition. In this work, we study continual and compositional personalization in VAR models and propose C2P-VAR, a unified framework addressing both challenges. For continual personalization, we introduce C2PVAR-S, which identifies concept-relevant parameters from gradient magnitudes and dynamically updates their selection during training. To preserve previously learned concepts, C2PVAR-S applies regularization only to parameters shared by the current and historical concepts, thereby reducing unnecessary interference without introducing additional model components. For multi-concept personalization, we further propose C2PVAR-M, which employs parallel global and concept-specific branches with spatially localized feature fusion and logit aggregation to achieve controllable concept placement and reduce feature entanglement. Extensive experiments on continual and multi-concept personalization demonstrate that C2P-VAR consistently outperforms existing baselines in subject fidelity, while maintaining competitive text alignment and introducing negligible storage and inference overhead. Our results establish a unified framework for scalable and controllable personalization of visual autoregressive models.
\end{abstract}

\section{Introduction}

Recent advances in text-to-image generation 
\citep{peebles2023scalable,ramesh2022hierarchical,rombach2022high,xu2023imagereward,sun2026flowerase} have been largely driven by diffusion models, which achieve remarkable visual quality through iterative denoising processes. More recently, Visual Autoregressive (VAR) models \citep{tian2024visual,han2025infinity} have emerged as a promising alternative, reformulating image generation as a coarse-to-fine next-scale prediction problem. By shifting the autoregressive unit from spatial tokens to resolution scales, VAR enables both high-quality synthesis and significantly improved inference efficiency \citep{zhong2025closing}.

Despite these advantages, current VAR models remain limited in personalized generation. In practical applications, users often wish to generate images involving 
specific concepts, such as personal objects or customized styles, which are difficult to fully describe via text prompts alone. Existing personalization approaches address this by fine-tuning models on a small set of images, achieving promising results for single-concept generation. However, real-world personalization is inherently dynamic and incremental. Users continuously introduce new concepts over time, necessitating a continual personalization framework. Unfortunately, existing 
methods \citep{zhao2024motiondirector,yang2025lora,chen2024anydoor} struggle in this setting. First, sequential learning of new concepts leads to catastrophic forgetting, where previously acquired concepts are 
overwritten \citep{rebuffi2017icarl,dong2022federated}. Second, composing multiple learned concepts in a single image introduces severe feature entanglement, 
resulting in incorrect attribute binding and visual artifacts 
\begin{wrapfigure}{r}{0.61\textwidth}
  \centering
  \includegraphics[width=0.6\textwidth]{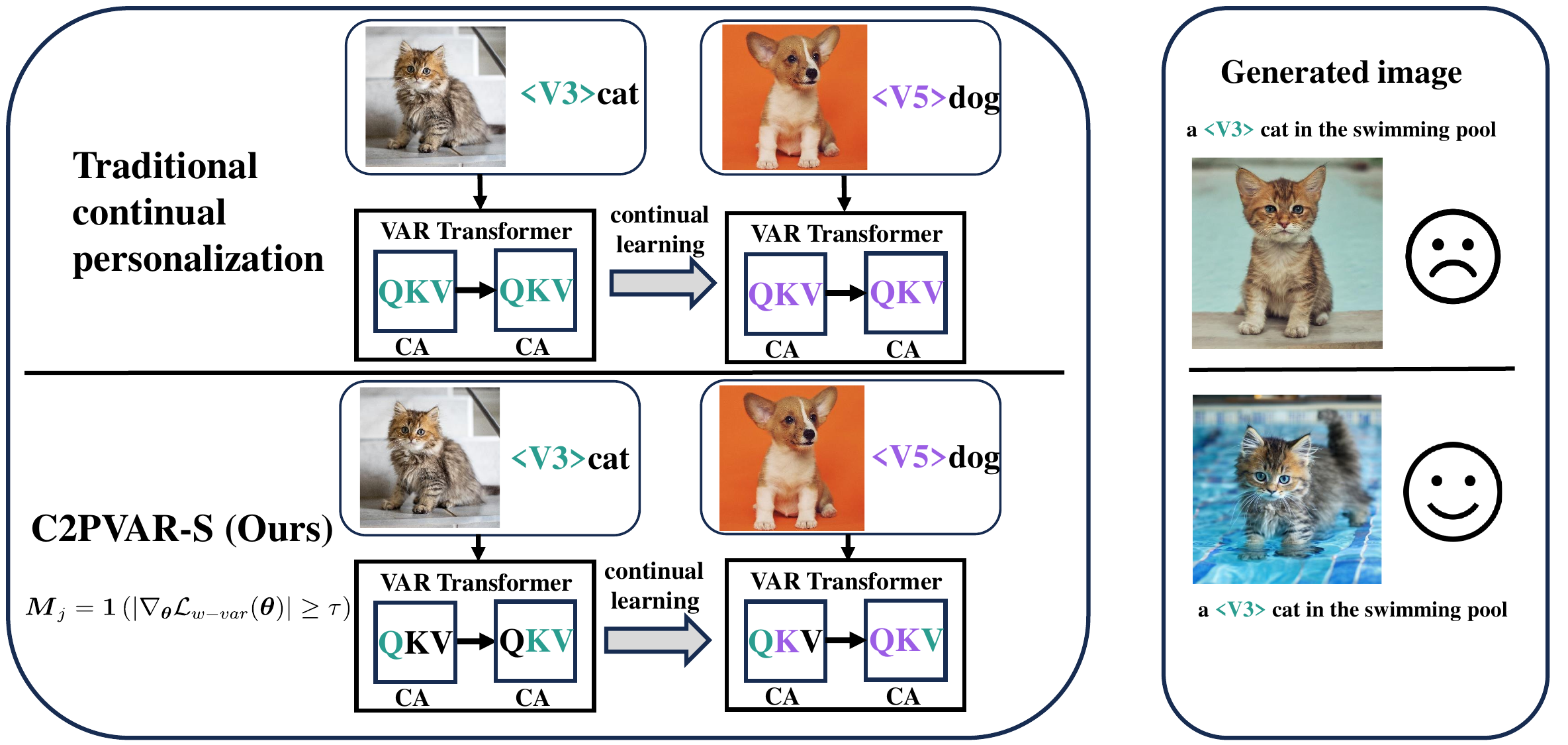}
  \caption{Schematic overview of our method (C2PVAR-S) versus the traditional full fine-tuning continual personalization approach: generating a special cat after learning two concepts.}
  \vspace{-1em}
  \label{fig:introjpg}
\end{wrapfigure}
\citep{chefer2023attend,feng2022training,jang2024identity,lee2023aligning,ma2024directed,wu2023human,yu2022scaling}. 
While continual learning and compositional generation have been extensively studied in diffusion models \citep{smith2023continual,kumari2023multi,po2024orthogonal}, directly transferring these techniques to VAR proves ineffective. Our empirical study reveals that existing methods suffer from substantial performance degradation when applied to VAR architectures, due to differences in generation dynamics and representation structure.

In this paper, we present the first systematic investigation of continual personalized generation in VAR models.
We identify two fundamental challenges: preserving concept knowledge across sequential updates and enabling controllable multi-concept composition. As illustrated in the Figure \ref{fig:introjpg}, traditional full-parameter fine-tuning during continual personalization causes subsequently learned concepts to completely overwrite previously acquired ones, thereby resulting in attribute overriding and catastrophic forgetting. To address these challenges, we propose a unified framework with two key components. For continual single-concept learning, we introduce C2PVAR-S, which dynamically identifies parameters most relevant to each concept based on gradient contributions. By enforcing regularization only on overlapping parameters across tasks, C2PVAR-S effectively mitigates catastrophic forgetting while avoiding unnecessary constraints on unrelated parameters. For multi-concept synthesis, we propose C2PVAR-M, which is tailored to the hierarchical generation process of VAR. Specifically, we perform multi-branch feature modeling at critical scales and introduce spatially guided cross-attention fusion using user-provided conditions such as bounding boxes \citep{li2023gligen,shuai2406survey}. This design enables precise control over concept placement and significantly reduces feature interference.

Extensive experiments demonstrate that our approach  outperforms existing baselines in both long-horizon continual personalization and multi-concept image synthesis. Our results highlight the unique challenges of personalization in VAR models and establish a strong foundation for future research in scalable and controllable generative systems. Our contributions are summarized as follows:

\begin{itemize}
    \item We present the first study of continual personalized generation in VAR models, revealing the limitations of existing approaches in this setting.
    \item We propose C2PVAR-S, a parameter-efficient method that mitigates catastrophic forgetting via concept-specific parameter selection and conflict-aware regularization. We introduce C2PVAR-M, that enables precise and disentangled composition within the VAR framework.
    \item We demonstrate strong empirical performance across both continual learning and compositional generation benchmarks.
\end{itemize}

\section{Related work}
\paragraph{Personalization within the VAR framework}
Concept personalization \citep{chen2024anydoor,gal2023encoder,kim2024selectively,li2023blip,motamed2023lego,zhang2024attention} aims to adapt pre-trained T2I generation models to synthesize personalized concepts utilizing only a limited number of exemplar images. Common personalized concepts typically encompass specific subjects or distinct visual styles. Pioneering work, such as ARBooth \citep{chung2025fine}, achieved single-concept personalization within the VAR architecture for the first time through a Selective Layer Tuning strategy. This approach assigns a unique identifier \citep{ruiz2023dreambooth} to a user-specific concept and specifically finetunes the Feed-Forward Network (FFN) and Cross-Attention (CA) layers of the VAR model. However, this method inherently assumes that the user trains the model on merely a single specific concept. Consequently, it falls short of satisfying the practical demand for continual concept learning and is fundamentally incapable of facilitating multi-concept generation.

\paragraph{Continual concept personalization}
Continual concept personalization further aims to incrementally expand the repertoire of concepts learned within a T2I model. A variety of representative works have already been established based on diffusion models \citep{gu2023mix,kumari2023multi}.  \citet{smith2023continual} proposed a self-regularization loss among the LoRA weights of distinct tasks to preserve previously acquired concepts. However, as the number of concepts increases, the newly introduced LoRA weights become heavily constrained by all previously learned LoRA weights, resulting in a substantial degradation in both the learning capacity for new concepts and the robustness against catastrophic forgetting. \citet{po2024orthogonal} proposes to represent the weight residual of each concept using a fixed orthogonal basis $B_i$ and a trainable low-rank matrix $A_i$, while encouraging the residual subspaces of different concepts to be approximately orthogonal. However, as the number of concepts increases, residual interference may still arise, potentially leading to catastrophic forgetting.

\section{Method}\label{sec4}

\subsection{Preliminary}
Under VAR \citep{tian2024visual}, a vision encoder maps an image \(x\) to a continuous feature map \(F \in \mathbb{R}^{h \times w \times c}\). The quantizer \(Q\) progressively converts \(F\) into discrete token maps \(\{r_s\}_{s=1}^{S}\) at \(S\) spatial resolutions. At scale \(s\), the residual feature is
\begin{equation}
f_s = \sum_{i=1}^{s}\mathrm{up}(r_i,(h,w)),
\end{equation}
where \(\mathrm{up}(\cdot,\cdot)\) upsamples a single-scale feature map to the target resolution, and \(f_s\) aggregates all scales up to \(s\). During inference,
\begin{equation}
\tilde{f}_s = \mathrm{down}(f_s,(h_{s+1},w_{s+1}))
\end{equation}
is prepended as the initial tokens for predicting the next scale. A scale-wise causal mask enables localized bidirectional information flow, while the transformer predicts the residual feature map at the subsequent scale. Infinity \citep{han2025infinity} adapts VAR for text-to-image generation by replacing the original VQ quantizer \citep{van2017neural} with BSQ \citep{zhao2024image}. Conditioned on a text prompt \(c\), its likelihood is
\begin{equation}
p(r_1,r_2,\dots,r_S)
= \prod_{s=1}^{S} p(r_s \mid r_1,r_2,\dots,r_{s-1};c).
\end{equation}

\subsection{Problem formulation and framework overview}
For Custom Visual Autoregressive Model (CVAR), given \(N\) subject images 
\(X=\{x_n\}_{n=1}^{N}\) and a text prompt \(c_{\mathrm{sub}}\) containing a concept token (e.g., \(\langle S^* \mathrm{dog} \rangle\)), personalization minimizes
\begin{equation}
\mathcal{L}_{\mathrm{var}}
= -\sum_{s=1}^{S}\log p_{\theta}
(r_s\mid r_1,r_2,\dots,r_{s-1};c_{\mathrm{sub}}),
\label{eq:lvar}
\end{equation}
where \(r_s\) denotes the multi-scale token maps extracted from \(x_n\). Most CVARs assume a fixed number of personalized concepts, limiting their ability to accommodate evolving user demands. C2P-VAR therefore decomposes continual concept generation into (i) preserving previously learned concepts during sequential customization and (ii) controllably composing multiple personalized concepts.

\begin{figure}[t]
	\centering
	\includegraphics[width = 1.0\textwidth]{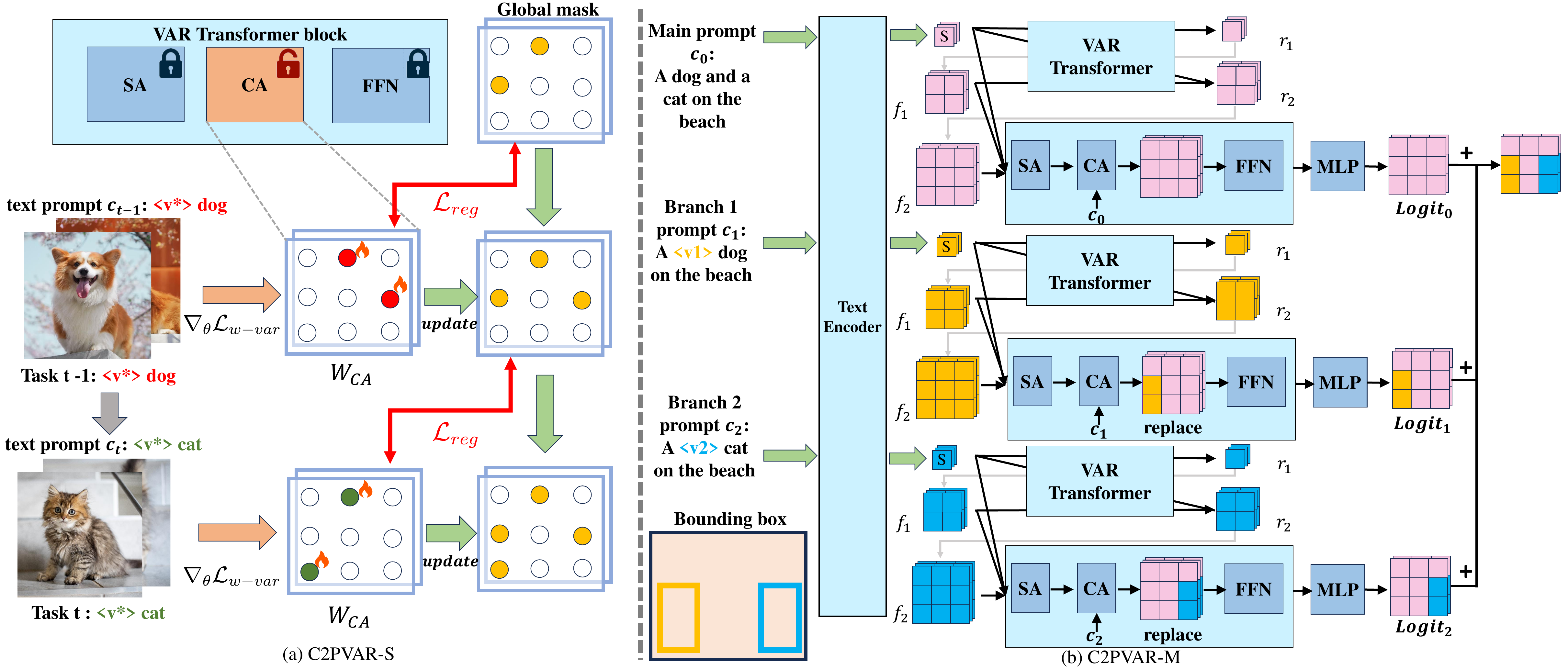}
	\caption{\textbf{Overall framework.} (a) C2PVAR-S resolves catastrophic forgetting and enable continual personalization. (b) C2PVAR-M addresses the challenge of concept neglect in multi-concept image synthesis.}
	\label{fig:pipeline}
\end{figure}

To mitigate catastrophic forgetting, C2PVAR-S selects a compact set of concept-specific parameters according to absolute gradient magnitude; only these parameters are updated and regularized, as described in Sec. \ref{sec:42}. C2PVAR-M performs multi-branch feature modeling and spatially guided localized cross-attention fusion for precise concept composition, as described in Sec. \ref{sec:43}. The overall framework is shown in Figure \ref{fig:pipeline}.

\subsection{Single-concept continual learning}\label{sec:42}

\textbf{Concept parameter selection. }
When full-parameter fine-tuning is adopted, sequentially learning an increasing number of concepts on a single model leads to severe interference: updates for newly introduced concepts tend to overwrite parameters associated with previously learned ones, resulting in catastrophic forgetting. Therefore, it is crucial to develop a parameter selection mechanism that minimizes overlap across concepts and preserves previously acquired knowledge.

Our goal is to identify a mechanism that directs concept personalization toward a subset of model parameters that are most relevant to the target concept, thereby reducing parameter overlap across different concepts. Inspired by gradient-based saliency methods \citep{smilkov2017smoothgrad,fan2023salun} for input attribution, we extend this idea to the parameter space and ask: can we construct a weight saliency map to guide continual concept personalization? This perspective allows us to partition model parameters into two subsets: (i) salient parameters that are critical for representing the current concept and should be updated, and (ii) non-salient parameters that remain frozen to preserve previously learned knowledge. Empirically, prior work such as ARBooth \citep{chung2025fine} has shown that cross-attention layers are strongly correlated with personalization objectives. In contrast, we observe that updating feed-forward network (FFN) layers often introduces interference among similar concepts. Based on these observations, we restrict parameter selection to the cross-attention layers of the VAR model.

Formally, consider the $t$-th task with model parameters $\theta_{t} \in \mathbb{R}^D$. During each training epoch, we compute the gradient $g \in \mathbb{R}^D$ of the loss function with respect to the parameters in the cross-attention layers:
\begin{equation}
    g = \nabla_{\theta_t}\mathcal{L}_{var}(\theta_t),
\end{equation}
where $\mathcal{L}_{var}(\theta_t)$ refers to Equation \ref{eq:lvar} defined previously. We define parameters whose absolute gradients exceed the threshold as the essential parameters for the current task (see details in the supplementary material).  We then construct a binary importance mask $M \in \{0, 1\}^D$ for the model. For the $j$-th parameter $\theta_j$ within the model, the activation rule for its corresponding mask $m_j$ is formulated as follows:
\begin{equation}
   m_j = \begin{cases} 1, & \text{if } |g_j| \ge \tau \\ 0, & \text{otherwise} \end{cases} .
\end{equation}
When $m_j = 1$, it indicates that the given parameter is strongly activated by the current concept and must be included in the protected list.

\textbf{Dynamic mask updating. }
We observe that important parameters vary across different training stages, and using a fixed mask increases interference among similar concepts, leading to catastrophic forgetting (Table \ref{tab:ablation_components}). To address this, we periodically refresh the mask during training. For task $t$, the gradient distribution is recalculated every $e$ epochs to generate a phase-specific mask $M_t^{ke}$, where $k \in \mathbb{N}$. After training, all phase-specific masks are merged through a logical OR operation to obtain the final task mask $M_t$ during $E$ training epochs:
\begin{equation}
    M_{t} = \bigvee_{k=0}^{\lfloor E/e \rfloor} M^{ke}_{t}.
\end{equation}
\textbf{Cross-task conflict regularization. }
Although task-specific masks identify important parameters, they cannot prevent interference across tasks. When learning task $t$, updates for the new concept may overwrite parameters important to previous tasks. To mitigate this, we introduce a global mask-based regularization mechanism. Before training task $t$, we first aggregate the masks of all previous tasks to form a historical mask $M_{<t}$:
\begin{equation}
    M_{<t} = \bigvee_{i=1}^{t-1} M_i.
\end{equation}
During training, parameters with $M_{<t,j}=0$ are freely updated, while overlapping parameters with $M_{<t,j}=1$ are constrained by an $L_2$ regularization term. The overall objective is:
\begin{equation}
    \mathcal{L}_{total} = \mathcal{L}_{var}(\theta_t) + \lambda \left\| M_{reg} \odot (\theta_t - \theta_{old}) \right\|_2^2,
\label{eq:ltotal}
\end{equation}
where $\theta_{old}$ denotes the model weights of the previous task, $\lambda$ is the regularization coefficient, $M_{reg}=M_{<t} \land M_t^{ke}$ denotes the overlap between the current update mask and historical mask, and $\odot$ denotes the Hadamard product.

\textbf{Scale-wise weighted loss. }
We observe that coarse scales have a larger impact on generation quality than fine scales in VAR. Therefore, we prioritize learning at coarse scales by introducing a scale-weighted cross-entropy loss $\mathcal{L}_{w-var}$ to replace $\mathcal{L}_{var}$:
\begin{equation}
\mathcal{L}_{w-var} = -\sum_{s=1}^{S} w_s \log p_{\theta}(r_s \mid r_{1},r_{2},\dots,r_{s-1}; c_{\mathrm{sub}}).
\label{eq:lwvar}
\end{equation}
\subsection{Spatially Guided Concept Composition
}\label{sec:43}
When generating scenes containing multiple objects, traditional CVAR models often suffer from feature confusion (unintended blending between objects) and subject neglect (only one object appears). To address this, we propose C2PVAR-M for multi-concept generation. 
Given $B$ customized concepts, the inference condition consists of one global condition and $B$ local conditions. The global condition $y_{global}$, describing the overall scene, is assigned to the $0$-th branch to establish the image layout. Each local condition contains a prompt $y_i$ and a bounding box $b_i$, where $i \in \{1,\dots,B\}$. The prompt $y_i$ includes the special token learned during single-concept training and is assigned to the corresponding local branch.

We observe that when the resolution scale reaches $s \ge 3$, the global spatial structure is largely determined, making this stage suitable for spatial intervention. During inference at $s \ge 3$, the global and local branches first independently fuse text features through cross-attention. To constrain each concept within its target region while preserving global consistency elsewhere, we replace local features outside the target mask with global features.

Let $f_G \in \mathbb{R}^{L_q \times d}$ denote the global branch feature map at the cross-attention layer, and $f_i \in \mathbb{R}^{L_q \times d}$ denote the feature map of the $i$-th local branch. Let $\mathbf{1}$ be an all-ones tensor with the same shape as $b_i$. The fused local feature is computed as:
\begin{equation}
    f^{F}_{i} = b_{i} \odot f_{i} + (\mathbf{1} - b_{i}) \odot f_{G}.
\label{eq:Ffused}
\end{equation}
This operation preserves local concept features inside the mask while inheriting global features outside it. To better integrate personalized concepts with the background, we further perform logits-level fusion. Let $L_G$ and $L_i$ denote the predicted logits of the global branch and the $i$-th local branch, respectively. We introduce a hyperparameter $\alpha$ (set to $0.05$ in our experiments) to control the influence of global features in local regions. The smoothed local logits are computed as:
\begin{equation}
    \tilde{L}_{i} = \alpha L_{G} + (1 - \alpha) L_{i}.
\end{equation}
We define the background mask as $b_G = \mathbf{1} - \bigvee_{i=1}^{B} b_i$. The final merged logits are then obtained by:
\begin{equation}
    L_{M} = b_{G} \odot L_{G} + \sum_{i=1}^{B} \left( b_{i} \odot \tilde{L}_{i} \right).
\label{eq:lmerged}
\end{equation}
Finally, the merged logits $L_M$ are synchronized across all branches for prediction at the next scale.

\section{Experiments}

\subsection{Experimental setups}\label{sec5.1}

\textbf{Datasets. } We construct the first challenging benchmark for concept incremental learning in VAR, consisting of 15 sequential concept customization tasks. Within this dataset, 13 customization tasks pertain to distinct object concepts collected from \citet{chung2025fine}, while the remaining two tasks involve different style concepts collected from websites. We establish approximately 3 to 5 text-image pairs for each task. Notably, we incorporate several semantically similar concepts (e.g., the dogs in V1 and V5), thereby rendering the dataset substantially more challenging within a continual personalization setting. More details are provided in the Appendix \ref{datasets}.

\textbf{Baselines. }
We compare our proposed method against two representative continual learning methods, three typical diffusion-based continual personalization approaches, and two standard finetuning baselines, which include LWF \citep{li2017learning}, MAS \citep{aljundi2018memory}, Continual Diffusion \citep{smith2023continual}, Orthogonal Adaptation \citep{po2024orthogonal}, CNS \citep{liao2025continual}, ARBooth \citep{chung2025fine}, and LoRA \citep{hu2022lora}. Further details are provided in the Appendix  \ref{baseline}.

\textbf{Evaluation metrics. }
Following \citet{gal2022image,kumari2023multi,nam2024dreammatcher,ruiz2023dreambooth}, we evaluate both subject fidelity and text prompt fidelity. To assess subject fidelity, we employ DINO \citep{caron2021emerging} and CLIP \citep{radford2021learning} to measure the image similarity between the generated images and the reference subjects, denoted as DINO and CLIP-I, respectively. To evaluate text prompt alignment, we compute the CLIP image-text similarity by comparing the visual features of the generated images with the textual features of the prompts (substituting the special token with its corresponding class token), denoting this metric as CLIP-T. Additionally, for each generated multi-concept image, we first crop the region corresponding to each concept according to the bounding boxes specified during the generation process. We then compute the DINO and CLIP-I between each cropped region and the corresponding reference image, and finally average the scores across all concepts. CLIP-T is still computed between the complete generated image and the multi-concept prompt to evaluate the overall text-image consistency.

\begin{figure}[t]
	\centering
	\includegraphics[width = 0.9\textwidth]{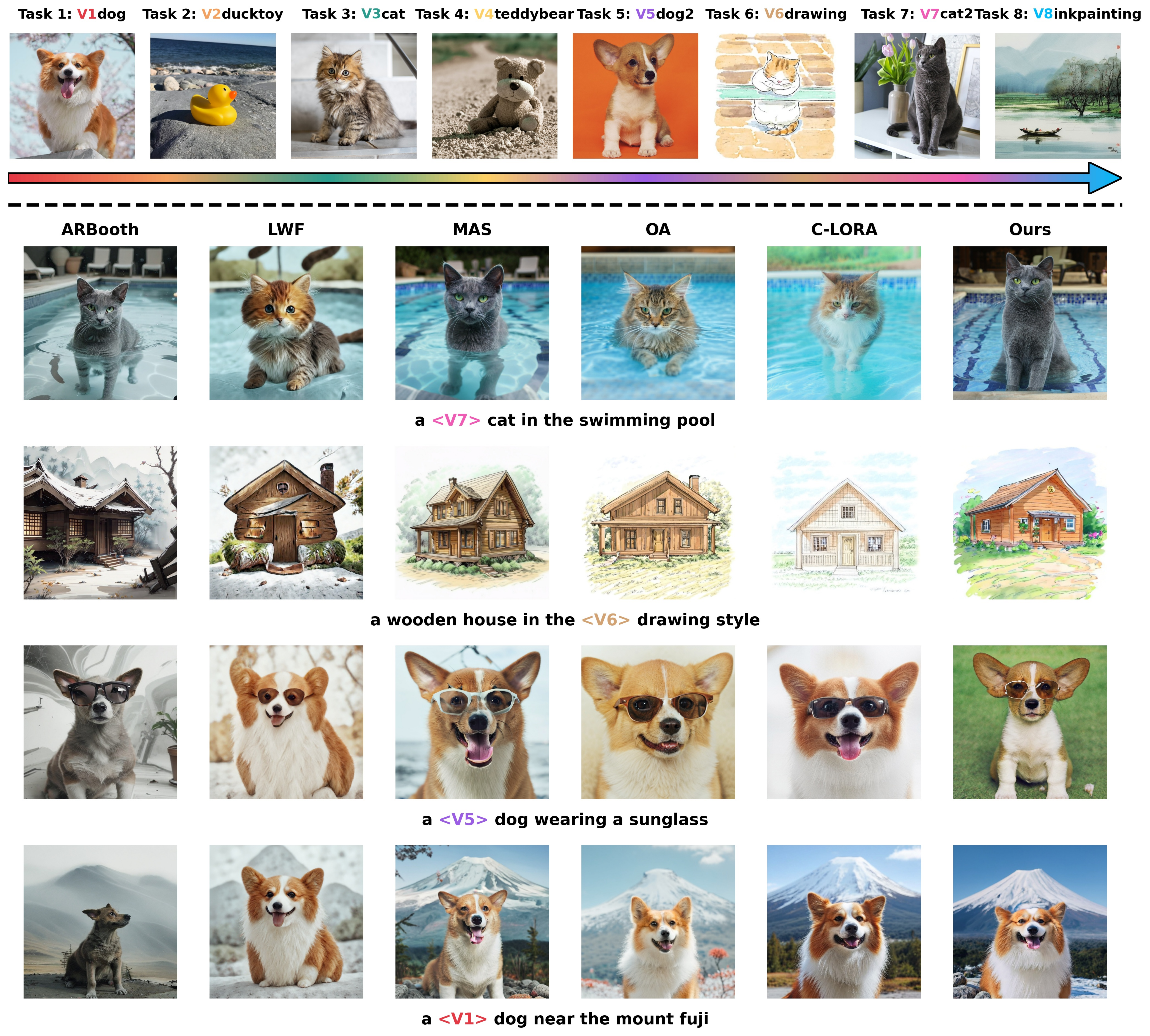}
	\caption{Qualitative comparison of single-concept customization with different baselines.}
	\label{fig:singleconcept}
\end{figure}

\subsection{Qualitative comparisons}
\begin{figure}[t]
	\centering
	\includegraphics[width = 0.9\textwidth]{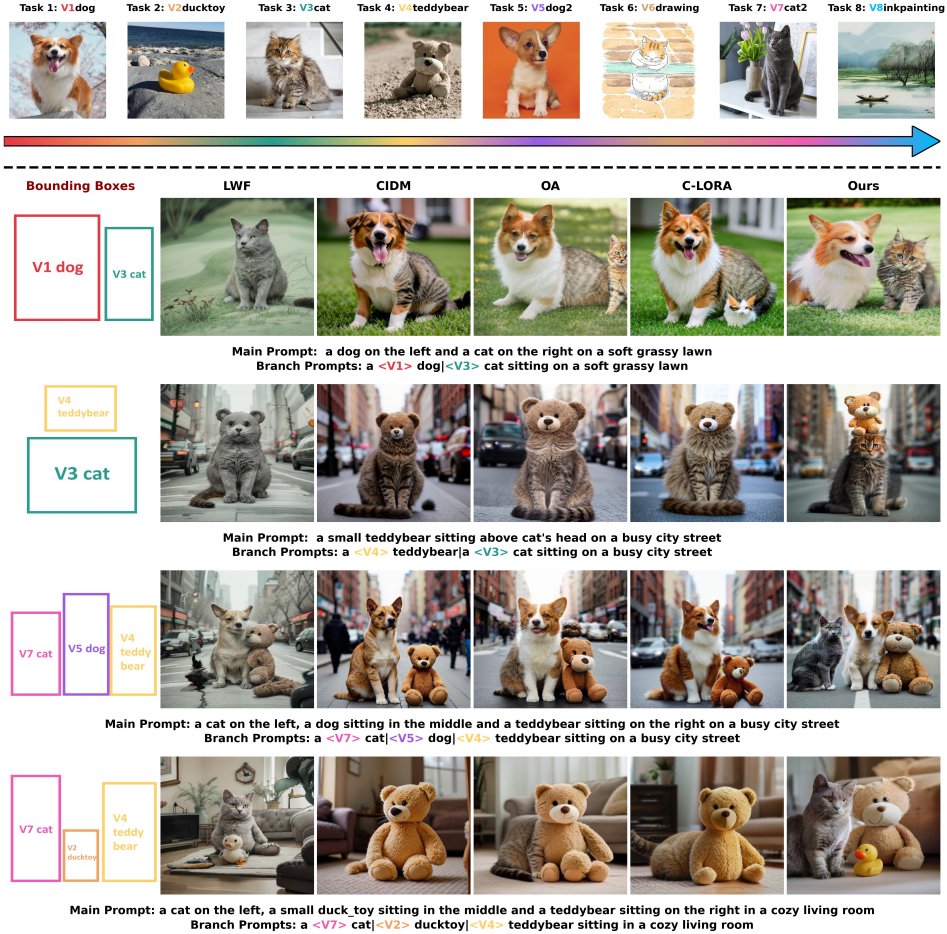}
	\caption{Qualitative comparison of multi-concept customization,
where Main Prompt indicates the global text prompt, and Branch Prompt denotes the region text prompt.}
	\label{fig:multiconcept}
\end{figure}

To validate the effectiveness of our model under concept incremental learning, we conduct qualitative comparisons on single-concept customization, multi-concept customization, and custom style transfer. As shown in Figure \ref{fig:singleconcept}, our model achieves superior single-concept customization by mitigating catastrophic forgetting while preserving the unique attributes of previously learned concepts. For multi-concept customization, we incorporate the region-controllable sampling module from \citet{gu2023mix} into baseline methods for fair comparison. As shown in Figure \ref{fig:multiconcept}, due to the architectural differences between diffusion models and VAR, prompts in VAR influence the generation process from the beginning rather than only through cross-attention layers. Therefore, directly applying the CA-based control strategy of \citet{gu2023mix} leads to severe concept neglect, where the generated image is dominated by the main prompt and fails to synthesize the target concepts. In contrast, our model effectively resolves this issue through C2PVAR-M. We provide more qualitative and quantitative results in Appendix.

\subsection{Quantitative comparisons}\label{sec5.3}

\begin{table}[h]
\centering
\caption{Quantitative comparisons of single-concept customization across 8 tasks. We report the DINO metric for each individual concept, alongside the average DINO, CLIP-I, and CLIP-T metrics.}
\vspace{1em}
\label{tab:single_concept_metrics_comprehensive}
\resizebox{\textwidth}{!}{
\begin{tabular}{lcccccccc|ccc}
\toprule
\multirow{2}{*}{\textbf{Methods}} & \multicolumn{8}{c|}{\textbf{Single Concept (DINO$\uparrow$)}} & \multicolumn{3}{c}{\textbf{Overall Metrics}} \\
\cmidrule(lr){2-9} \cmidrule(lr){10-12}
& \textbf{V1} & \textbf{V2} & \textbf{V3} & \textbf{V4} & \textbf{V5} & \textbf{V6} & \textbf{V7} & \textbf{V8} & \textbf{Avg. DINO$\uparrow$} & \textbf{Avg. CLIP-I$\uparrow$} & \textbf{Avg. CLIP-T$\uparrow$} \\
\midrule
ARBooth            & 40.78 & 45.71 & 62.36 & 66.03 & 61.62 & 17.80 & 78.97 & 40.16          & 51.68 & 75.99 & 31.28 \\
LoRA               & 40.32 & 33.40 & 73.80 & 64.52 & 58.78 & 19.76 & 73.68 & 40.56 & 50.60 & 76.63 & 31.48 \\
LWF                & 74.02 &	69.10 &	84.03 &	68.94 &	63.90 &	24.23 &	59.09 &	29.09 &	59.05 &80.78 &29.01 
 \\
MAS               & 66.88 &	59.94 &	57.05 &	65.77 	&74.55 &	29.73 &	74.97& 	28.83 &	57.21 &
77.52 & \textbf{32.65}

\\            
Orthogonal Adaptation & 79.06 &	57.52 	&80.17 &	70.40 &	67.23 	&22.12 &	60.68 	&27.21 &	58.05& 78.88& 31.58 

 \\
Continual Diffusion   & 82.49 &	57.59 &	83.82 &	73.44& 	69.28& 	23.30& 	63.66& 	26.78& 	60.04& 78.68& 31.85 

 \\
CNS   & 37.16 &	47.83& 	70.78 &	63.36 &	35.30& 	22.15 &	60.13 	&25.73 &	45.31 &75.61 &32.58 

 \\
\midrule
\textbf{C2PVAR-S (Ours)}  & 81.89&69.17 &82.86&75.50&80.36&
41.80&78.21&35.57& \textbf{68.17}& \textbf{82.71}& 31.14
\\
\bottomrule
\end{tabular}
}
\end{table}

We conduct quantitative comparisons exclusively on single-concept inference. For each concept, we evaluate it using 20 prompts and generate 10 images per prompt, resulting in a total of 200 images, over which the metrics are averaged. As presented in Table \ref{tab:single_concept_metrics_comprehensive}, we report the results of the first 8 concepts in the main paper for clarity, while the results for all 15 concepts are provided in the Appendix. Our method outperforms diffusion-based baselines in terms of both DINO and CLIP-I, demonstrating enhanced subject alignment. The results in both the main table and the Appendix further demonstrate that our method effectively mitigates catastrophic forgetting in both short-horizon and long-horizon continual learning settings, achieving the best overall performance. We additionally report the standard deviations of the three evaluation metrics across the 8 tasks in Appendix \ref{Additional quantitative comparisons}.

It is worth noting that the CLIP-T metric places a stronger emphasis on the overall alignment between the image background and the text prompt. However, our task definition inherently prioritizes the quality and fidelity of the customized subject. 
\begin{wraptable}{r}{0.6\textwidth}
\centering
\vspace{-1em}
\caption{Quantitative comparisons of multi-concept personalization. We report the average DINO, CLIP-I, and CLIP-T.}
\vspace{0.5em}
\label{tab:multi-concept_personalization}

\resizebox{0.48\textwidth}{!}{
\begin{tabular}{lccc}
\toprule
\textbf{Methods} 
& \textbf{Avg. DINO$\uparrow$} 
& \textbf{Avg. CLIP-I$\uparrow$} 
& \textbf{Avg. CLIP-T$\uparrow$} \\
\midrule
ARBooth               &  42.34&	73.15	&31.95\\
LoRA                  & 41.91&	74.66	&32.51 \\
LWF                   & 53.51&	78.89&	29.82 \\
MAS                   & 57.14	&79.12	&33.83 \\
Orthogonal Adaptation & 56.52	&80.44	&32.77 \\
Continual Diffusion   & 52.61&	79.36&	33.03 \\
CNS                   &  44.55	&79.16&	34.12\\
\midrule
\textbf{C2PVAR-M (Ours)}  & \textbf{68.37} & \textbf{83.35} & \textbf{35.24} \\
\bottomrule
\vspace{-2em}
\end{tabular}
}
\end{wraptable}
Consequently, there is typically a trade-off between the subject alignment metrics (DINO and CLIP-I) and the text alignment metric (CLIP-T). Taking this trade-off into 
consideration, our proposed method achieves the optimal overall performance.

In addition, we construct 45 multi-concept prompts, create an appropriate bounding box for each concept, and generate 10 images for each prompt, with the average results reported for evaluation. All prompts follow one controlled evaluation protocol. As shown in Table \ref{tab:multi-concept_personalization}, our method achieves significant improvements across all three metrics, demonstrating the best performance in both subject fidelity and text alignment for multi-concept personalization.


Concurrently, as shown in Appendix Table \ref{tab:computational_resources}, we evaluate the memory and time consumption of various methods when integrating eight distinct concepts. LoRA-based methods generally require additional memory and computational time because multiple concept-specific weights must be stored and composed during inference. In contrast, C2PVAR-S requires zero additional storage space and completely avoids the inference-time overhead associated with LoRA-weight composition. This design provides a more efficient deployment profile while maintaining high generative performance, thereby achieving an optimal balance between fusion efficiency and generation quality.

\section{Ablation study}

\begin{wrapfigure}{r}{0.61\textwidth}
  \centering
  \vspace{-1em}
  \includegraphics[width=0.6\textwidth]{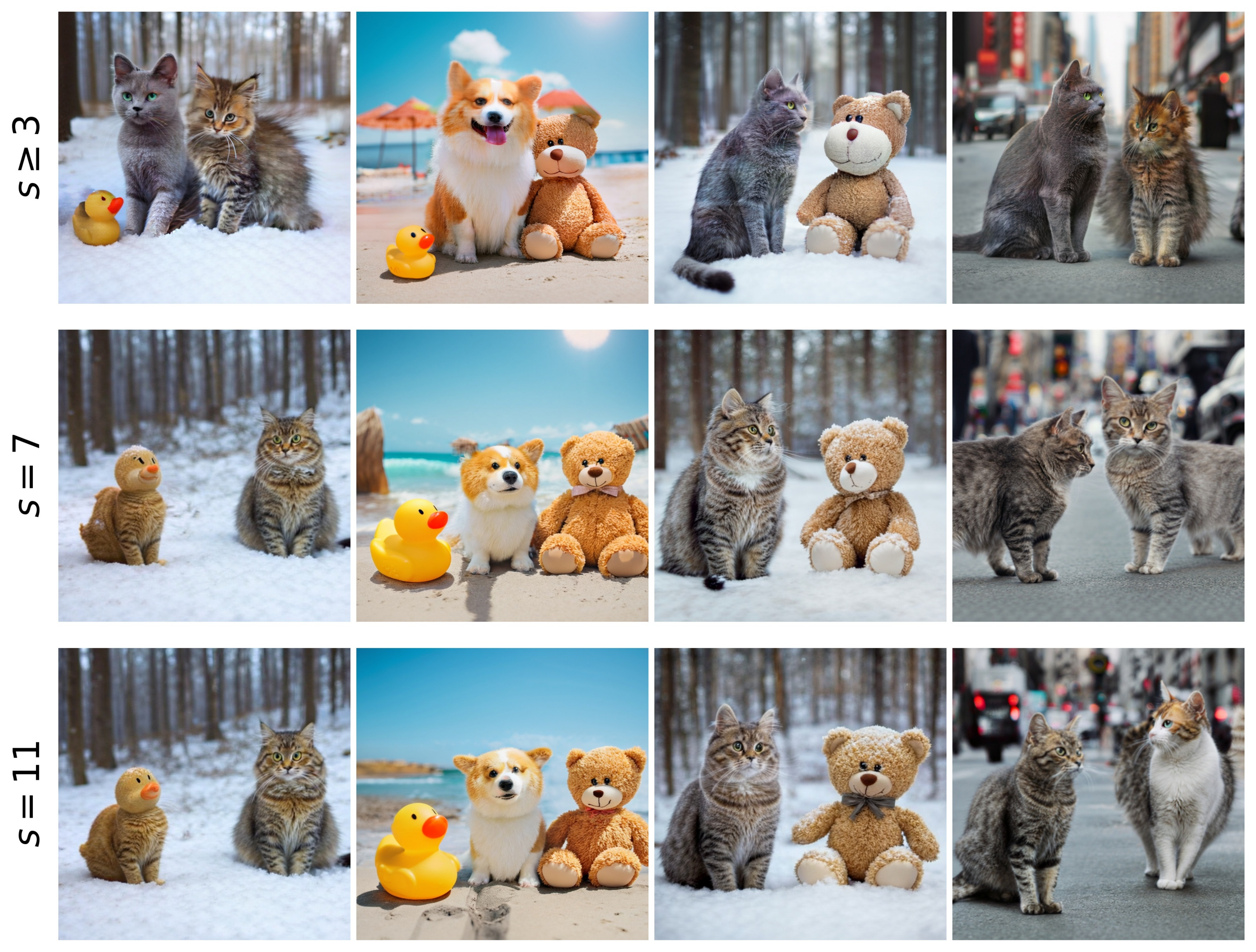}
	\caption{Ablation study on the intervention scale $s$}
	\label{fig:Ablation_Study_Levels}
\end{wrapfigure}

\paragraph{Impact of layer selection in multi-concept synthesis.}
To validate the necessity of our continuous multi-scale intervention (operating consistently from $s \ge 3$), we conducted an ablation study by applying the regional mask intervention only at a single high-resolution scale (e.g., exclusively at $s=7$ or $s=11$). Figure \ref{fig:Ablation_Study_Levels} shows that isolated intervention fails to synthesize the specific concepts and leads to severe image distortion, with the cross-attention exhibiting a negligible effect. Furthermore, owing to the sparse token count at lower resolution scales ($s=1$ and $s=2$), the bounding boxes are inevitably downscaled to encompass merely one or two tokens. This extreme spatial compression causes the features injected by the final processing branch to completely override the features established by all preceding branches, thereby rendering the method ineffective.

\paragraph{Effectiveness of regularization, dynamic masking and scale weighting.}
We investigate the effects of regularization, dynamic mask updating and scale weighting. 
As shown in Table \ref{tab:ablation_components}, without regularization loss, a subset of these overlapping parameters is overwritten by newly introduced concepts, consequently leading to catastrophic forgetting and a degradation in overall performance. We set $w_{>8} = 0.5$ and set the other weight to 1.0. As shown in the third row of the table, scale weighting effectively increases the DINO metrics, indicating an enhancement in the model's learning capacity. Meanwhile, dynamic mask updating effectively mitigates mutual interference among semantically similar concepts.  The DINO scores for the V1dog and V3cat exhibit substantial improvements, demonstrating the successful alleviation of catastrophic forgetting.

\begin{table*}[tbp]
\centering
\caption{Ablation study on the effectiveness of Regularization, Scale Weighting, and Dynamic Masking. We report the DINO scores for each concept and the overall average.}
\vspace{1em}
\label{tab:ablation_components}
\resizebox{\textwidth}{!}{
\begin{tabular}{ccc|ccccccccc}
\toprule
\multicolumn{3}{c|}{\textbf{Components}} & \multicolumn{9}{c}{\textbf{Single Concept(DINO $\uparrow$})} \\
\cmidrule(lr){1-3} \cmidrule(lr){4-12}
\textbf{Reg.} & \textbf{Weight} & \textbf{Mask} & \textbf{V1} & \textbf{V2} & \textbf{V3} & \textbf{V4} & \textbf{V5} & \textbf{V6} & \textbf{V7} & \textbf{V8} & \textbf{Avg.} \\
\midrule
           &            &            &58.80 &	56.10 &	73.98 	&64.77 	&64.93 	&21.21 	&65.37 &	29.41 	&54.32  \\
\checkmark &            &           &  71.76	&60.58	&79.84&	77.29&	81.37	&33.32&	78.17&	33.96&	64.54
\\

\checkmark & \checkmark &    & 74.74&	63.52&	76.72	&76.08&	80.36&	38.31&	78.75&	33.79	&65.28
 \\

\checkmark & \checkmark & \checkmark & 	81.89&69.17 &82.86&75.50&80.36&
41.80&78.21&35.57& \textbf{68.17}
\\
\bottomrule
\end{tabular}
}
\end{table*}


  
%
  
%
  
%
  
%

\section{Conclusion}\label{sec7}

We address two challenges in VAR-based personalization: catastrophic forgetting during concept-incremental customization and feature entanglement in multi-concept synthesis. We present C2P-VAR, the first continual and compositional personalization framework for VAR. C2PVAR-S updates task-relevant parameter subspaces with conflict-aware regularization, mitigating forgetting without data replay. C2PVAR-M enables spatially controlled feature fusion and branch-wise logit aggregation during inference. Experiments show that C2P-VAR outperforms state-of-the-art diffusion methods in both tasks while maintaining low computational and storage overhead.

\section*{AI use statement}

In this work, we used generative AI tools solely for language polishing of the manuscript, including improving the clarity and readability of existing text. We did not use generative AI tools to generate or develop research ideas, methodology, experiments, analysis, or scientific claims. All AI-assisted text was carefully reviewed and verified by the authors, who take full responsibility for the final content of this work.

\section*{Ethics statement}
Our work does not involve human subjects, human participant studies, or the collection of personally identifiable information. All concept images used in our experiments are obtained from publicly available datasets or publicly accessible sources, and are used solely for research and evaluation purposes in accordance with the usage conditions and licenses of their respective sources. We do not collect, process, or release any private or sensitive personal data.

Our work investigates continual learning and multi-concept personalization for generative models. While generative models may have potential dual-use risks, we are not aware of any specific harmful applications introduced by the proposed method beyond the general risks associated with image generation and personalization. We follow standard research practices and report our experimental results according to the stated evaluation protocols. We have no known conflicts of interest or external sponsorship that could influence this work.

\section*{Reproducibility statement}
All implementation details, including the hyperparameter settings and experimental configurations, are described in Appendix \ref{implement} and Appendix \ref{appendix a3}, providing the necessary details for reproducing our experiments.
The datasets used for experiments are already publicly available.

\bibliography{iclr2027_conference}
\bibliographystyle{iclr2027_conference}

\appendix

\section{Appendix}

\subsection{Baseline method}\label{baseline}

\paragraph{ARBooth \citep{chung2025fine}} We clone the code base of ARBooth from official GitHub repository. The learning rate of text embedding is 2e-5 and Infinity model are 2e-5 while training steps is set to 300. For multi-concept personalization, we naively feed all the trained special tokens we need to the model.

\paragraph{LoRA \citep{hu2022lora}} We naively integrate LoRA into the CA and FFN layers of the Infinity model, with the rank $r$ set to $64$, and conduct continual training on the same pair of LoRA matrices. The learning rate of text embedding is 2e-5 and LoRA matrices are 2e-5 while training steps is set to 300. For multi-concept personalization, we naively feed all the trained special tokens we need to the model.

\paragraph{LWF \citep{li2017learning}} 

We adapt LwF to the Infinity model. For each new concept, the student model is initialized from the checkpoint of the previous task, while a frozen copy of the same checkpoint is used as the teacher model. During training, we align the autoregressive output distributions of the teacher and the currently optimized student model with a temperature-scaled KL divergence loss, following the standard LwF objective:
\begin{equation}
\mathcal{L}
=
\mathcal{L}_{\mathrm{new}}
+
\lambda_{\mathrm{LwF}}
\tau^2
\sum_{t}
\mathrm{KL}
\left(
\mathrm{softmax}\left(\frac{z^{T}_{t}}{\tau}\right)
\;\middle\|\;
\mathrm{softmax}\left(\frac{z^{S}_{t}}{\tau}\right)
\right),
\end{equation}
where \(z^{T}_{t}\) and \(z^{S}_{t}\) denote the teacher and student autoregressive logits at token position \(t\), and \(\tau\) is the distillation temperature. The learning rates of both the text embedding and the Infinity model are set to 2e-5, and each concept is trained for 300 steps. For multi-concept personalization, we naively feed all the trained special tokens required by the prompt to the model.

\paragraph{MAS \citep{aljundi2018memory}} 
We adapt MAS to the Infinity model as an importance-regularization baseline.
Before optimizing the new concept, we load the accumulated MAS importance weights
from previous tasks and penalize changes to important historical parameters with
a quadratic regularization term as follows:
\begin{equation}
\mathcal{R}_{\mathrm{MAS}}(\theta)
=
\lambda_{\mathrm{MAS}}
\sum_i \Omega_i \left(\theta_i - \theta_i^{*}\right)^2 ,
\end{equation}
where \(\Omega_i\) denotes the accumulated importance of parameter \(\theta_i\),
and \(\theta_i^{*}\) denotes its value after learning previous tasks.
After finishing each task, we estimate the current-task importance by computing
the gradient magnitude of the squared autoregressive logits with respect to the
trainable parameters, and merge it into the historical importance with a decay
factor. The learning rates of both the text embedding and the Infinity model are
set to 2e-5, and each concept is trained for 300 steps. For
multi-concept personalization, we naively feed all the trained special tokens we
need to the model.

\paragraph{Orthogonal Adaptation \citep{po2024orthogonal}}
We adapt Orthogonal Adaptation to the Infinity model by ourselves because official code is not available. For each adapted linear layer, the weight residual is parameterized as $\Delta W_i = A_i B_i^T$, where $B_i$ is fixed and sampled from a shared randomized orthogonal basis, while only $A_i$ and the concept embedding are optimized. We apply the adapters to the cross-attention and feed-forward linear layers, train each concept independently for 300 steps, and set the learning rates of both the text embedding and adaptation parameters to 2e-5. During inference, multiple concepts are merged by summing their learned residual weights with a fixed merge coefficient.

\paragraph{Continual Diffusion \citep{smith2023continual}}
We adapt Continual Diffusion to the Infinity model by ourselves, as the official code is unavailable.
We follow the self-regularization loss presented in Continual Diffusion to fulfill continual personalization.
For the $t$-th concept, we follow the C-LoRA self-regularization objective in Continual Diffusion.
Let $\Delta W_t^\ell$ denote the LoRA-induced weight update of layer $\ell$ for the current concept,
and let $\Delta W_{<t}^\ell = \sum_{i=1}^{t-1} \Delta W_i^\ell$ denote the accumulated LoRA updates from previous concepts.
The self-regularization loss is formulated as below:
\begin{equation}
\mathcal{L}_{\mathrm{C\text{-}LoRA}}
=
\sum_{\ell \in \mathcal{A}}
\left\|
\Delta W_{<t}^\ell \odot \Delta W_t^\ell
\right\|_F^2,
\end{equation}
where $\odot$ denotes element-wise multiplication and $\mathcal{A}$ is the set of adapted layers.
We apply the adapters to the cross-attention and feed-forward linear layers.
The learning rates of both the text embedding and LoRA parameters are set to 2e-5, and each concept is trained for 300 steps.

\paragraph{CNS \citep{liao2025continual}}
We adapt CNS to the Infinity model by ourselves, as the official code is unavailable. For each new concept, we train only the selected rows of the $W_k$ and $W_v$ matrices in cross-attention layers. Following CNS, given a caption embedding c and a key/value weight matrix $W$, we compute the neuron importance as $S(W,c)=|W|\odot(\mathbf{1}\cdot\|c\|_2)$ denotes the channel-wise text-embedding intensity and is broadcast to all rows of $W_k$ and $W_v$. We use ImageNet calibration captions to identify general neurons and the current concept captions to identify concept-specific neurons. We set the row selection ratio to 5\% and use $\lambda_{1}$=1.0 and $\lambda_{2}$=1.0 for the overlap regularization terms following the paper setting. The learning rates of both the text embedding and the Infinity model are set to 2e-5, and each concept is trained for 300 steps. For multi-concept personalization, we naively feed all the trained special tokens we need to the model.

\subsection{Implementation details}
\label{implement}We conduct all experiments utilizing the Infinity-2B model \citep{han2025infinity}, which is pretrained on the LAION \citep{schuhmann2021laion}, COYO \citep{kakaobrain2022coyo-700m}, and OpenImages \citep{kuznetsova2020open} datasets. We adopt the default scale configuration ($S = 13$). We employ the AdamW optimizer \citep{loshchilov2017decoupled} ($\beta_0 = 0.9, \beta_1 = 0.97$), setting the learning rate to 2e-5 for the text embeddings and 2e-5 for the concept neurons. The model is finetuned for 300 iterations at a resolution of 1024 with a batch size of 1 on a single NVIDIA A6000 GPU. We empirically set $\lambda$ to 1, and detailed ablation experiments are presented in Table \ref{tab:lambda_ablation}.

\subsection{Datasets}\label{datasets}

We collect 13 object concepts from \citet{chung2025fine} and 2 style concepts from online sources, resulting in a total of 15 concepts: V1 dog, V2 duck toy, V3 cat, V4 teddy bear, V5 dog, V6 style, V7 cat, V8 style, V9 clock, V10 dog, V11 monster toy, V12 sunglasses, V13 teapot, V14 vase, and V15 wooden barn. For multi-concept image evaluation, we further construct 45 prompts with corresponding bounding boxes, including 25 prompts featuring two object concepts arranged vertically, 15 prompts featuring two object concepts arranged horizontally, and 5 prompts involving three concepts. Each multi-concept prompt specifies the target concepts and their spatial arrangement, together with the corresponding bounding boxes used to control their locations during generation. An example of the constructed multi-concept prompts and their corresponding bounding boxes is provided in the Table \ref{tab:multiconcept_prompt_example}. We evaluate our framework and all competing baselines under the same dataset and evaluation settings.

\begin{table}[h]
\centering
\caption{An example of the multi-concept prompts and corresponding bounding boxes used in our evaluation.}
\vspace{1em}
\label{tab:multiconcept_prompt_example}
\resizebox{\textwidth}{!}{
\begin{tabular}{p{0.28\textwidth}|p{0.60\textwidth}}
\toprule
\textbf{Item} & \textbf{Example} \\
\midrule
Task 
& V1 dog + V2 duck toy  \\

Branch prompts 
& ``a V1 dog sitting in a magical garden''; 
``a V2 duck toy sitting in a magical garden'' \\

Main prompt 
& ``A photo of a dog sitting on the left and a duck toy sitting on the right in a magical garden'' \\

Bounding boxes 
& dog: $[0.1, 0.25, 0.45, 0.8]$; 
duck toy: $[0.65, 0.25, 0.95, 0.8]$ \\
\bottomrule
\end{tabular}
}
\end{table}

\subsection{Threshold of the neuron selection}\label{appendix a3}
We identify the top 5\% of parameters as the crucial concept parameters for each task. We conduct an additional ablation study on the threshold for selecting crucial concept parameters. As shown in the Table \ref{tab:threshold_ablation}, a threshold of 5\% achieves the best trade-off among the three evaluation metrics, demonstrating the most balanced overall performance.

\begin{table}[htbp]
\centering
\caption{Ablation study on the threshold for concept parameters selection. We report the average DINO, CLIP-I, and CLIP-T scores across all tasks.}
\vspace{1em}
\label{tab:threshold_ablation}
\begin{tabular}{lccc}
\toprule
\textbf{Threshold} & \textbf{Avg. DINO} $\uparrow$ & \textbf{Avg. CLIP-I} $\uparrow$ & \textbf{Avg. CLIP-T} $\uparrow$ \\
\midrule
5\%  & 68.17& 82.71& 31.14 \\
1\%  & 66.15 &81.69& 31.31 \\
10\% & 67.65 &83.02& 30.69 \\
\bottomrule
\end{tabular}
\end{table}

\subsection{Additional quantitative comparisons}
\label{Additional quantitative comparisons}
We have added additional quantitative  comparisons in this section. As shown in Table \ref{tab:single_concept_clipi_tasks}, Table \ref{tab:single_concept_clipt_tasks},
our method outperforms diffusion-based
baselines in terms of CLIP-I, demonstrating enhanced subject alignment. It is worth
noting that the CLIP-T metric places a stronger emphasis on the overall alignment between the image
background and the text prompt. However, our task definition inherently prioritizes the quality and
fidelity of the customized subject. Consequently, there is typically a trade-off between the subject
alignment metrics (DINO and CLIP-I) and the text alignment metric (CLIP-T). 
As shown in Table \ref{tab:single_concept_dino_tasks_15}, Table \ref{tab:single_concept_clipi_tasks_15} and \ref{tab:single_concept_clipt_tasks_15},
our method continues to outperform all baselines in terms of both DINO and CLIP-I in the long-horizon continual learning setting, demonstrating its strong robustness and effectiveness under prolonged continual learning. As shown in Table \ref{tab:single-concept-statics}, 
we additionally report the standard deviations of the three evaluation metrics across the 8 tasks, obtained from three independent runs with different random seeds. These results further demonstrate the reliability of our evaluation and the robustness of our method.

As shown in Table \ref{tab:computational_resources},
we evaluate the memory and time consumption of various methods
when integrating eight distinct concepts. It is noteworthy that LoRA-based methods necessitate memory and computational time resources
that substantially exceed those required by C2PVAR-S. In contrast, C2PVAR-S requires zero additional storage space and completely circumvents the inference-time overhead typically associated with LoRA
weights composition, thereby demonstrating an optimal balance between fusion efficiency and high
generative performance.

\begin{table}[h]
\centering
\caption{Quantitative comparisons of single-concept customization across 8 tasks (CLIP-I metric).}
\vspace{1em}
\label{tab:single_concept_clipi_tasks}

\resizebox{\textwidth}{!}{
\begin{tabular}{lccccccccc}
\toprule
\multirow{2}{*}{\textbf{Methods}} & \multicolumn{9}{c}{\textbf{Single Concept (CLIP-I$\uparrow$)}} \\
\cmidrule(lr){2-10}
& \textbf{V1} & \textbf{V2} & \textbf{V3} & \textbf{V4} & \textbf{V5} & \textbf{V6} & \textbf{V7} & \textbf{V8} & \textbf{Avg.} \\
\midrule
ARBooth               &  76.26 & 	77.43 & 	77.67 & 	79.51 	& 77.48&  	61.68 & 	80.87 & 	77.03 & 	75.99 
\\
LoRA                   & 79.67 & 	72.80 & 	81.46 & 	81.90 & 	78.88&  	61.36 	& 80.02 	& 76.94 	& 76.63 
 \\
LWF                    & 86.37 &	84.18 &	89.21 &	88.64 &	85.87 &	62.25 &	79.22 &	70.52 &	80.78 
 \\  
MAS                    & 84.33 &	82.49 &	77.53 	&81.81& 	82.35 &	60.74 &	80.92 &	70.02 &	77.52 
 \\   
Orthogonal Adaptation  & 86.15 &	83.46 &	85.78 &	84.30 &	83.59 	&62.68 	&75.82 &	69.30 &	78.88 
 \\
Continual Diffusion    & 87.79 &	82.95 &	83.66 &	84.91 	&83.50 &	62.07 &	76.66 &	67.91 &	78.68 
\\
CNS    & 79.20& 	81.00 &	79.24 &	84.07& 	76.10 &	60.47 	&76.80 	&68.00 &	75.61 
\\
\midrule
\textbf{C2PVAR-S (Ours)}    & 85.75&84.83&86.04&89.87&87.32&67.13&84.67&76.06&\textbf{82.71 } 
\\
\bottomrule
\end{tabular}
}
\end{table}

\begin{table}[h]
\centering
\caption{Quantitative comparisons of single-concept customization across 8 tasks (CLIP-T metric).}
\vspace{1em}
\label{tab:single_concept_clipt_tasks}

\resizebox{\textwidth}{!}{
\begin{tabular}{lccccccccc}
\toprule
\multirow{2}{*}{\textbf{Methods}} & \multicolumn{9}{c}{\textbf{Single Concept (CLIP-T$\uparrow$)}} \\
\cmidrule(lr){2-10}
& \textbf{V1} & \textbf{V2} & \textbf{V3} & \textbf{V4} & \textbf{V5} & \textbf{V6} & \textbf{V7} & \textbf{V8} & \textbf{Avg.} \\
\midrule
ARBooth                & 30.52 &	33.02& 	30.54 &	32.53 &	30.59 &	31.33 &	30.75 &	30.96 	&31.28  \\
LoRA                   & 30.08 &	32.41& 	30.85& 	33.06 	&30.86& 	31.76 &	31.91& 	30.92 &	31.48 \\
LWF                    & 26.09 &	31.96 	&28.26 &	29.24 &	26.92 &	30.58 &	28.83 &	30.17 &	29.01 
 \\   
 MAS                    & 30.69 &	33.92 &	31.86 &	34.10 &	30.71 &	34.06 	&31.83 	&34.01 &	32.65 
 \\  
Orthogonal Adaptation  & 28.70 &	34.24 &	29.33 &	32.23 &	29.84 &	33.43 &	32.32 &	32.51 &	31.58 
 \\
Continual Diffusion    & 30.05 &	33.02 &	31.74 	&32.19 &	30.07& 	33.29 	&32.36 &	32.08 &	31.85 
 \\
CNS    & 31.26 &	33.34 &	32.22 &	33.17 	&31.30 &	34.17 &	32.28 &	32.92 	&32.58 
 \\
\midrule
\textbf{C2PVAR-S (Ours)}    & 31.26& 33.72&29.85&31.41&29.53&32.33&29.71&31.31&31.14 \\
\bottomrule
\end{tabular}
}
\end{table}

\begin{table}[h]
\centering
\caption{Quantitative comparisons of single-concept customization across 15 tasks (DINO metric).}
\vspace{1em}
\label{tab:single_concept_dino_tasks_15}

\resizebox{\textwidth}{!}{
\begin{tabular}{lcccccccccccccccc}
\toprule
\multirow{2}{*}{\textbf{Methods}} 
& \multicolumn{16}{c}{\textbf{Single Concept (DINO$\uparrow$)}} \\
\cmidrule(lr){2-17}
& \textbf{V1} & \textbf{V2} & \textbf{V3} & \textbf{V4} & \textbf{V5} 
& \textbf{V6} & \textbf{V7} & \textbf{V8} & \textbf{V9} & \textbf{V10} 
& \textbf{V11} & \textbf{V12} & \textbf{V13} & \textbf{V14} & \textbf{V15}
& \textbf{Avg.} \\
\midrule
ARBooth                & 44.73 &	44.62 &	60.64 &	65.24 &	27.09 &	18.32 &	58.54 	&28.48 &	58.64& 	62.70 &	42.78 &	53.98 &	54.45 &	56.60 &	84.98 &	50.79 
 \\
LoRA                   & 31.59 &	34.88 &	57.33 &	63.21 &	27.35 &	20.08 &	48.25 &	32.84 &	58.85 &	41.73 	&42.63 &	56.30 &	52.20 	&51.08 &	82.15 &	46.70 
 \\
LWF                    & 67.01 &	65.32 &	78.75 &	55.83 &	64.13 &	25.72 &	64.44 &	20.21 	&36.18 	&21.19 &	27.71 &	58.06 &	46.01& 	47.12 &	60.93 	&49.24 
 \\
MAS                    & 65.09 &	64.72& 	62.24 &	67.66& 	41.38 	&33.31& 	75.20& 	31.79 &	69.26 &	72.70 &	26.87 &	67.35 &	48.81 &	48.44 &	72.01 &	56.46 
 \\
Orthogonal Adaptation  & 70.64 &	54.95 &	74.98 &	63.90 &	54.32& 	20.21 	&60.70 &	21.80 &	49.63 &	48.53 &	23.84 &	53.00 &	45.00 &	29.63 &	44.22 &	47.69 
 \\
Continual Diffusion    & 76.59 &	50.32 &	80.51 &	72.00 &	59.62 &	19.42 &	62.67 &	28.01 &	68.45 &	46.97 &	36.49 &	64.14 &	54.50 &	37.01 &	56.23 	&54.19 
 \\
CNS                    & 36.40 &	48.01 	&71.13 &	62.57& 	34.99 &	22.75 	&60.04 &	25.49 &	59.40 	&26.10& 	28.15 &	59.82 &	48.72 &	37.92 &	67.87&	45.96
 \\
\midrule
\textbf{C2PVAR-S (Ours)}   & 84.06& 63.50& 87.11& 75.76& 80.71& 40.80& 78.93& 37.45& 71.25& 80.14& 56.04& 64.52& 67.14& 67.36 &78.44& \textbf{68.88} 
 \\
\bottomrule
\end{tabular}
}
\end{table}

\begin{table}[h]
\centering
\caption{Quantitative comparisons of single-concept customization across 15 tasks (CLIP-I metric).}
\vspace{1em}
\label{tab:single_concept_clipi_tasks_15}

\resizebox{\textwidth}{!}{
\begin{tabular}{lcccccccccccccccc}
\toprule
\multirow{2}{*}{\textbf{Methods}} 
& \multicolumn{16}{c}{\textbf{Single Concept (CLIP-I$\uparrow$)}} \\
\cmidrule(lr){2-17}
& \textbf{V1} & \textbf{V2} & \textbf{V3} & \textbf{V4} & \textbf{V5} 
& \textbf{V6} & \textbf{V7} & \textbf{V8} & \textbf{V9} & \textbf{V10} 
& \textbf{V11} & \textbf{V12} & \textbf{V13} & \textbf{V14} & \textbf{V15}
& \textbf{Avg.} \\
\midrule
ARBooth & 74.74 & 79.63 & 74.70 & 77.95 & 70.08 & 55.87 & 72.11 & 68.15 & 77.03 & 78.54 & 68.19 & 78.46 & 73.50 & 76.79 & 86.25 & 74.13 
 \\
LoRA & 73.28 & 76.04 & 76.95 & 82.50 & 71.74 & 58.76 & 75.10 & 73.12 & 78.65 & 75.08 & 68.73 & 79.20 & 74.72 & 75.41 & 82.76 & 74.80 
 \\
LWF & 85.88 & 81.31 & 86.10 & 81.77 & 84.01 & 62.50 & 82.69 & 66.69 & 70.93 & 73.61 & 63.22 & 79.97 & 75.75 & 75.81 & 70.33 & 76.04 
 \\
MAS & 81.56 & 82.33 & 77.85 & 83.51 & 75.50 & 62.64 & 80.76 & 68.55 & 82.11 & 83.12 & 61.95 & 82.76 & 74.12 & 77.00 & 78.60 & 76.82 
 \\
Orthogonal Adaptation & 80.20 & 85.40 & 80.39 & 81.63 & 79.80 & 61.67 & 76.71 & 68.00 & 79.76 & 74.62 & 61.00 & 79.14 & 77.92 & 72.85 & 65.06 & 74.94 
 \\
Continual Diffusion & 85.82 & 81.05 & 81.34 & 83.19 & 80.01 & 58.97 & 76.53 & 67.98 & 81.65 & 77.56 & 60.93 & 80.95 & 77.52 & 74.57 & 67.42 & 75.70 
 \\
CNS & 78.93 & 80.84 & 79.20 & 83.81 & 75.88 & 60.73 & 76.59 & 67.91 & 81.17 & 78.30 & 62.71 & 80.18 & 73.79 & 72.99 & 77.02 & 75.34 
 \\
\midrule
\textbf{C2PVAR-S (Ours)} & 86.83&  82.57&  90.85&  88.65&  89.16&  66.56&  87.49&  77.04&  84.54&  84.98&  76.12 & 81.97&  84.65&  82.70&  80.14&  \textbf{82.95}
 \\
\bottomrule
\end{tabular}
}
\end{table}

\begin{table}[h]
\centering
\caption{Quantitative comparisons of single-concept customization across 15 tasks (CLIP-T metric).}
\vspace{1em}
\label{tab:single_concept_clipt_tasks_15}

\resizebox{\textwidth}{!}{
\begin{tabular}{lcccccccccccccccc}
\toprule
\multirow{2}{*}{\textbf{Methods}} 
& \multicolumn{16}{c}{\textbf{Single Concept (CLIP-T$\uparrow$)}} \\
\cmidrule(lr){2-17}
& \textbf{V1} & \textbf{V2} & \textbf{V3} & \textbf{V4} & \textbf{V5} 
& \textbf{V6} & \textbf{V7} & \textbf{V8} & \textbf{V9} & \textbf{V10} 
& \textbf{V11} & \textbf{V12} & \textbf{V13} & \textbf{V14} & \textbf{V15}
& \textbf{Avg.} \\
\midrule
ARBooth & 30.22 & 33.54 & 31.22 & 33.23 & 30.41 & 28.75 & 30.90 & 28.22 & 29.81 & 28.25 & 31.09 & 33.21 & 34.97 & 33.82 & 32.33 & 31.33 \\
LoRA & 29.83 & 33.26 & 32.31 & 32.77 & 31.23 & 31.43 & 32.46 & 30.71 & 30.91 & 30.21 & 31.32 & 34.85 & 35.42 & 34.13 & 32.81 & 32.24 \\
LWF & 26.01 & 30.87 & 27.55 & 27.89 & 26.81 & 29.06 & 27.15 & 28.01 & 24.28 & 25.13 & 24.41 & 30.25 & 30.77 & 27.31 & 30.83 & 27.76 \\
MAS & 30.56 & 34.39 & 31.41 & 34.00 & 30.64 & 33.02 & 31.61 & 32.95 & 29.48 & 29.97 & 31.63 & 33.24 & 37.90 & 35.58 & 33.42 & 32.65 \\
Orthogonal Adaptation & 27.67 & 31.72 & 29.13 & 32.39 & 29.05 & 33.10 & 30.15 & 30.44 & 28.60 & 28.29 & 28.33 & 30.07 & 33.29 & 30.74 & 30.25 & 30.22 \\
Continual Diffusion & 30.36 & 34.37 & 32.07 & 32.55 & 30.76 & 32.70 & 32.60 & 31.60 & 29.86 & 30.65 & 29.91 & 30.52 & 34.54 & 32.54 & 30.88 & 31.73 \\
CNS & 31.22 & 33.29 & 32.27 & 33.43 & 31.35 & 34.32 & 32.32 & 33.09 & 29.25 & 30.79 & 32.68 & 34.23 & 38.04 & 34.83 & 33.97 & 33.01 \\
\midrule
\textbf{C2PVAR-S (Ours)} & 31.04&  33.50 & 28.66&  31.99&  28.04&  32.44&  27.57&  32.06 & 27.98&  28.45 & 31.04&  31.31&  31.75&  30.74&  32.98&  30.64
\\
\bottomrule
\end{tabular}
}
\end{table}

\begin{table}[htbp]
\centering
\caption{Quantitative comparisons of single-concept customization across 8 tasks. We report the average DINO, CLIP-I, and CLIP-T metrics with standard deviations.}
\vspace{1em}
\label{tab:single-concept-statics}
\resizebox{0.6\textwidth}{!}{%
\begin{tabular}{lccc}
\toprule
\textbf{Methods} 
& \textbf{Avg. DINO$\uparrow$} 
& \textbf{Avg. CLIP-I$\uparrow$} 
& \textbf{Avg. CLIP-T$\uparrow$} \\
\midrule
ARBooth                   & 51.68 
 $\pm$ 0.82 & 75.99 
 $\pm$ 0.37 & 31.28 
 $\pm$ 0.14 \\
LoRA                   & 50.60 
 $\pm$ 0.65 & 76.63 
 $\pm$ 0.39 & 31.48 
 $\pm$ 0.53 \\
LWF                   & 59.05 
 $\pm$ 0.44 & 80.78 
 $\pm$ 0.80 & 29.01 
 $\pm$ 0.30 \\
MAS                   & 57.21 
 $\pm$ 0.38 & 77.52 
 $\pm$ 0.25 & \textbf{32.65 
 $\pm$ 0.15} \\
Orthogonal Adaptation & 58.05 
 $\pm$ 0.29 & 78.88 
 $\pm$ 0.18 & 31.58 
 $\pm$ 0.58 \\
Continual Diffusion   & 60.04 
 $\pm$ 0.19 & 78.68 
 $\pm$ 0.14 & 31.85 
 $\pm$ 0.07 \\
CNS                   & 45.31 
 $\pm$ 0.26 & 75.61 
 $\pm$ 0.58 & 32.58 
 $\pm$ 0.12\\
\midrule
\textbf{C2PVAR-S (Ours)}  & \textbf{68.17 
 $\pm$ 0.47} & \textbf{82.71 
 $\pm$ 0.07} & 31.14 
 $\pm$ 0.14 \\
\bottomrule
\end{tabular}%
}
\end{table}

\subsection{Additional qualitative  comparisons}

\begin{figure}[h]
	\centering
	\includegraphics[width = 0.6\textwidth]{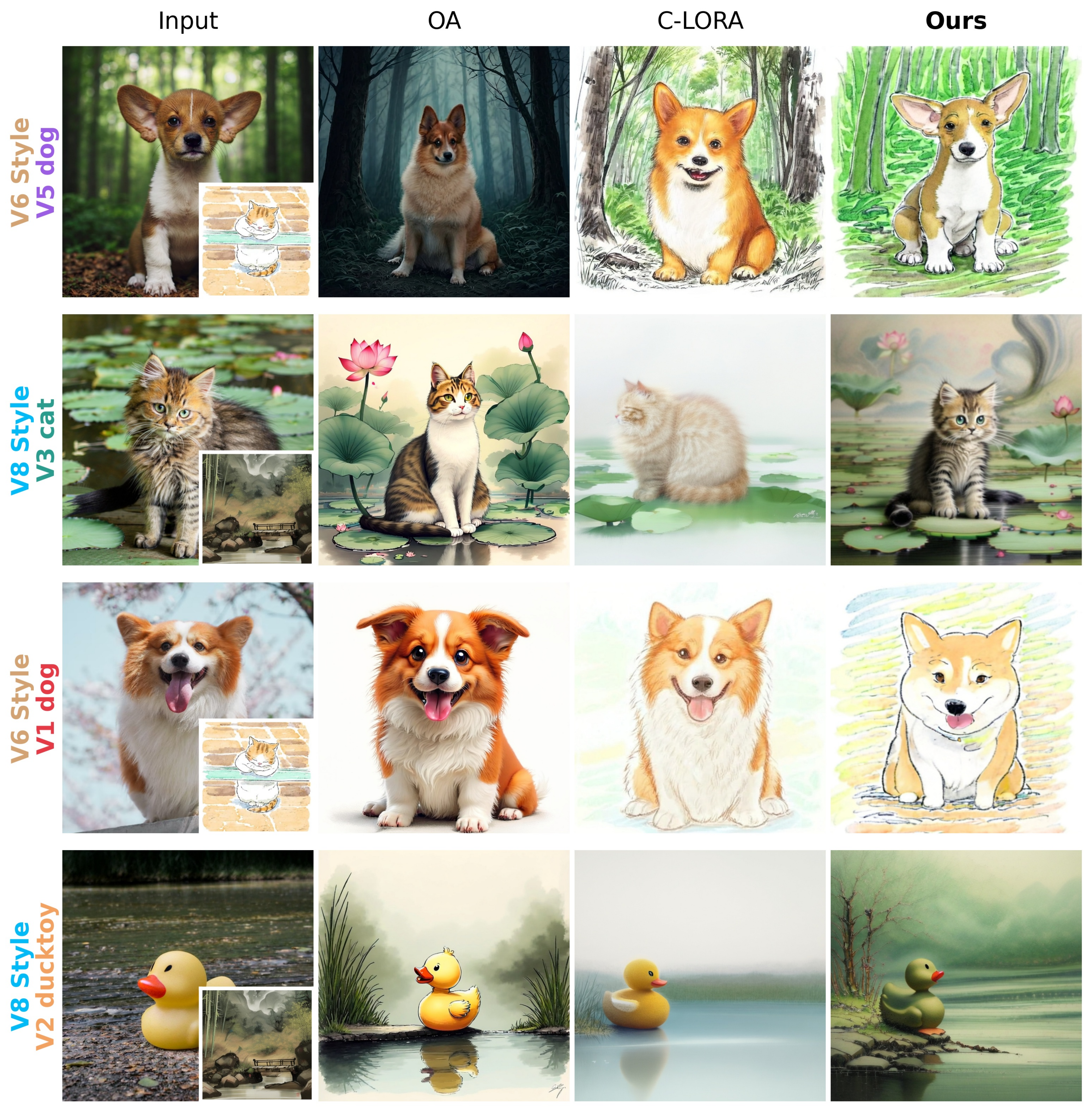}
	\caption{Qualitative comparison of custom style transfer}
	\label{fig:styleconcept}
\end{figure}

We have added additional qualitative  comparisons in this section, as shown in Figure \ref{fig:styleconcept}. Our proposed method inherently facilitates the seamless transfer of personalized style concepts to personalized object concepts. Moreover, the image generation process merely requires the inclusion of both the style special token and the object special token within the text prompt, entirely circumventing the need for any auxiliary multi-concept synthesis strategies. This capability arises because distinct concept tokens can directly activate their respective 
concept parameters within the model, thereby enabling the generation of the specific object accurately rendered in the designated style.

\subsection{Pseudocode of the framework}

In this section, we provide the detailed pseudocode for our proposed framework. Algorithm \ref{alg:gcns} outlines the continual fine-tuning process with C2PVAR-S, which fundamentally resolves catastrophic forgetting. Algorithm \ref{alg:multi_concept} details the multi-concept synthesis inference process.

\begin{algorithm}[H]
\caption{C2PVAR-S}
\label{alg:gcns}
\begin{algorithmic}[1]

\STATE \textbf{Input:} Pre-trained Infinity model weights $\theta$, Task datasets $\mathcal{T}=\{\mathcal{D}_1, \dots, \mathcal{D}_T\}$ 
\STATE \textbf{Hyper-parameters:} learning rates $\eta$, selection ratio $p$ , update interval $e$, penalty $\lambda$ , iterations $N$.

\vspace{0.2cm}
\STATE Initialize $M_{<t} \leftarrow \mathbf{0}$, $\theta_{old} \leftarrow \theta$ 
\FOR{each task $t = 1$ to $T$}
    \STATE $M_{t} \leftarrow \mathbf{0}$, $k \leftarrow 0$
    \FOR{epoch $ep = 0$ to $E$}
        \IF{$ep \bmod e == 0$}
            \STATE $g \leftarrow \nabla_{\theta}\mathcal{L}_{w-var}(\theta; \mathcal{D}_t)$ \COMMENT{Compute gradients of CA layers }
            \STATE $M_t^{ke}\leftarrow|g| \ge \text{Percentile}(|g|, 100-p)$
            \STATE $M_{t} \leftarrow M_{t} \lor M_t^{ke}$ \COMMENT{Update task mask}
            \STATE $k\leftarrow k+1$
        \ENDIF
        \FOR{iteration $i = 1$ to $N$}
            
            \STATE $M_{reg} \leftarrow M_{<t} \land M_t^{ke}$ \COMMENT{Detect overlapping parameters }
            \STATE $\mathcal{L}_{total} \leftarrow \mathcal{L}_{w-var}(\theta) + \lambda \left\| M_{reg} \odot (\theta - \theta_{old}) \right\|_2^2$ \COMMENT{Eq.\ref{eq:ltotal}}
            \STATE $\theta \leftarrow \theta - \eta \cdot \nabla_{\theta}\mathcal{L}_{total}$

        \ENDFOR
    \ENDFOR
    \STATE $M_{<t} \leftarrow M_{<t} \lor M_{t}$, $\theta_{old} \leftarrow \theta$ \COMMENT{Update global mask}
\ENDFOR
\RETURN $\theta_T$
\end{algorithmic}
\end{algorithm}

\begin{algorithm}[H]
\caption{C2PVAR-M}
\label{alg:multi_concept}
\begin{algorithmic}[1]
\STATE \textbf{Input:} Global prompt $y_{global}$, local prompts and bboxes $\{(y_{i}, b_{i})\}_{i=1}^{B}$, Scales $S$
\STATE \textbf{Hyper-parameters:} Blending factor $\alpha$, intervention threshold $s_{start}$.
\vspace{0.2cm}
\STATE Initialize $B+1$ parallel branches and compute background mask $b_{G} \leftarrow \mathbf{1} -\bigvee_{i=1}^{B} b_{i}$ 
\FOR{each scale $s = 1$ to $S$}
    \IF{$s \ge s_{start}$}
        \FOR{each VAR transformer block}
            \STATE $f_{G} \leftarrow \text{transformer}(y_{global})$
            \FOR{each local branch $i = 1$ to $B$}
                \STATE $f_{i} \leftarrow \text{transformer}(y_{i})$
                \STATE $f^{F}_{i} \leftarrow b_{i} \odot f_{i} + (\mathbf{1} - b_{i}) \odot f_{G}$ \COMMENT{Eq.\ref{eq:Ffused} }
            \ENDFOR
        \ENDFOR
        \STATE Obtain $L_{G}$ and $L_{i}$ after all VAR transformer blocks
        \STATE $L_{M} \leftarrow b_{G} \odot L_{G} + \sum_{i=1}^{B} \left( b_{i} \odot (\alpha L_{G} + (1-\alpha)L_{i}) \right)$ \COMMENT{Eq.\ref{eq:lmerged} }
        \STATE Synchronize all branches with $L_{M}$ 
        
    \ELSE
        \STATE Execute standard VAR next-scale prediction
    \ENDIF
    \STATE Sample tokens for scale $s$
\ENDFOR
\RETURN Final generated image $X$
\end{algorithmic}
\end{algorithm}

\begin{table}[htbp]
\centering
\caption{Ablation study on task order in continual learning. We additionally compare the average performance across eight concepts under three different learning sequences. Metrics are reported as the average scores over eight tasks.}
\vspace{1em}
\label{tab:order_ablation}
\begin{tabular}{lccc}
\toprule
\textbf{Task Order} & \textbf{Avg. DINO} $\uparrow$ & \textbf{Avg. CLIP-I} $\uparrow$ & \textbf{Avg. CLIP-T} $\uparrow$ \\
\midrule
Order 1 (Default) & 68.17 & 82.71 & 31.14 \\
Order 2 & 67.16 &81.40 &31.62 \\
Order 3 & 68.59& 83.14 &30.93 \\
Order 4 & 68.18& 82.53 &30.93 \\
\bottomrule
\end{tabular}
\end{table}

\begin{table}[htbp]
\centering
\caption{Ablation study on the cross-task conflict regularization coefficient $\lambda$. Metrics are reported as the average scores over eight tasks.}
\vspace{1em}
\label{tab:lambda_ablation}
\begin{tabular}{lccc}
\toprule
\textbf{Coefficient ($\lambda$)} & \textbf{Avg. DINO} $\uparrow$ & \textbf{Avg. CLIP-I} $\uparrow$ & \textbf{Avg. CLIP-T} $\uparrow$ \\
\midrule
$\lambda = 0.001$           &65.18 &80.46 &31.78  \\
$\lambda = 0.1$           &67.54 &82.68 &30.87  \\
$\lambda = 1.0$ (Default) & \textbf{68.17} &\textbf{82.71}  & 31.14 \\
$\lambda = 1000.0$           &67.67& 82.26 &\textbf{31.29}  \\
\bottomrule
\end{tabular}
\end{table}

\subsection{Additional ablation study}

\textbf{Sequence. } We also conduct an ablation study regarding the sequence of concepts in continual personalization. Specifically, order 2 is defined as (ducktoy, dog2, cat, drawing, teddybear, cat2, dog, inkpainting), order 3 as (ducktoy, drawing, teddybear, cat, dog2, cat2, inkpainting, dog), and order 4 as (dog, drawing, ducktoy, cat, dog2, inkpainting, teddybear, cat2). As presented in the Table \ref{tab:order_ablation}, our experiments fully demonstrate that alterations in the training sequence exert a negligible impact on the model's continual learning capabilities, thereby substantiating the strong robustness of our proposed C2PVAR-S continual learning method.

\textbf{Regularization coefficient. }
We evaluate the impact of varying $\lambda$ on the average performance across all customized concepts. $\lambda = 1.0$ is the default setting used in our framework, which achieves the optimal overall performance taking all three metrics into consideration.

\begin{table}[tbp]
\centering
\caption{Comparisons of computational resources. Memory requirements for GPU/CPU and computation time for customization indicate the additional costs for fusing concept weights previously learned.}
\vspace{1em}
\label{tab:computational_resources}
\begin{tabular}{lcc}
\toprule
\multirow{2}{*}{\textbf{Methods}} & \multicolumn{2}{c}{\textbf{Computational Resources}} \\
\cmidrule(lr){2-3}
& \textbf{Memory(MB)}$\downarrow$ & \textbf{Time(s)}$\downarrow$ \\
\midrule
ARBooth               & 0/0 & 0 \\
LoRA                  & 137/52 & 1.75 \\
LWF                   & 0/0 & 0 \\  
MAS                   & 0/0 & 0   \\
Orthogonal Adaptation & 1278/35 & 11.44 \\
Continual Diffusion   & 3303/60 & 12.51 \\
CNS                   & 0/0 & 0   \\
\midrule
\textbf{C2PVAR-S (Ours)}  & \textbf{0/0} & \textbf{0} \\
\bottomrule
\end{tabular}
\end{table}

\end{document}